\documentclass[letterpaper, 10 pt, conference]{ieeeconf}

\IEEEoverridecommandlockouts

\usepackage{cite}
\usepackage{ulem}
\usepackage{xcolor}
\usepackage{acronym}
\usepackage{caption}
\usepackage{leftidx}
\usepackage{listings}
\usepackage{graphicx}
\usepackage{textcomp}
\usepackage{multirow}
\usepackage{booktabs}
\usepackage{colortbl}
\usepackage{subcaption}
\usepackage{algorithmic}
\usepackage{amsmath,amssymb,amsfonts}

\acrodef{TP}{True Positive}
\acrodef{FP}{False Positive}
\acrodef{FN}{False Negative}
\acrodef{VSLAM}{Visual SLAM}
\acrodef{STD}{Standard Deviation}
\acrodef{ROS}{Robot Operating System}
\acrodef{RMSE}{Root Mean Square Error}
\acrodef{ATE}{Absolute Trajectory Error}
\acrodef{RANSAC}{RANdom SAmple Consensus}
\acrodef{CNN}{Convolutional Neural Network}
\acrodef{LiDAR}{Light Detection And Ranging}
\acrodef{SLAM}{Simultaneous Localization and Mapping}

\newcommand{\orb}{ORB-SLAM 3.0}

\newcommand{\etc}{\textit{etc. }}
\newcommand{\eg}{\textit{e.g., }}
\newcommand{\ie}{\textit{i.e., }}

\newcommand{\etal}{\textit{et al. }}

\definecolor{red}{HTML}{fd8f8f}
\definecolor{greend}{HTML}{57e377}
\definecolor{greenl}{HTML}{b8fb8a}
\definecolor{yellow}{HTML}{fefdb4}
\definecolor{orange}{HTML}{ffd5ab}

\colorlet{red}{red!50}
\colorlet{yellow}{yellow!50}
\colorlet{greenl}{greenl!50}
\colorlet{greend}{greend!50}
\colorlet{orange}{orange!50}

\title{\LARGE \bf Towards Localizing Structural Elements: Merging Geometrical Detection with Semantic Verification in RGB-D Data}

\author{
    Ali Tourani$^{1}$, Saad Ejaz$^{1}$, Hriday Bavle$^{1}$, Jose Luis Sanchez-Lopez$^{1}$, and Holger Voos$^{1}$
    \thanks{$^{1}$Authors are with the Automation and Robotics Research Group, Interdisciplinary Centre for Security, Reliability, and Trust (SnT), University of Luxembourg, Luxembourg. Ali Tourani is also associated with the Institute for Advanced Studies (IAS), University of Luxembourg, Luxembourg. Holger Voos is also associated with the Faculty of Science, Technology, and Medicine, University of Luxembourg, Luxembourg. \tt{\small{\{ali.tourani, saad.ejaz, hriday.bavle, joseluis.sanchezlopez, holger.voos\}}@uni.lu}}
    \thanks{*This research was funded, in whole or part, by the Luxembourg National Research Fund (FNR), DEUS Project, ref. C22/IS/17387634/DEUS. In addition, it was partially funded by the Institute of Advanced Studies (IAS) of the University of Luxembourg through an “Audacity” grant (project TRANSCEND - 2021).}
    \thanks{*For the purpose of open access, and in fulfillment of the obligations arising from the grant agreement, the author has applied a Creative Commons Attribution 4.0 International (CC BY 4.0) license to any  Author Accepted Manuscript version arising from this submission.}
}

\begin{document}

\maketitle
\thispagestyle{empty}
\pagestyle{empty}

\begin{abstract}
RGB-D cameras supply rich and dense visual and spatial information for various robotics tasks such as scene understanding, map reconstruction, and localization.
Integrating depth and visual information can aid robots in localization and element mapping, advancing applications like 3D scene graph generation and Visual Simultaneous Localization and Mapping (VSLAM).
While point cloud data containing such information is primarily used for enhanced scene understanding, exploiting their potential to capture and represent rich semantic information has yet to be adequately targeted.
This paper presents a real-time pipeline for localizing building components, including wall and ground surfaces, by integrating geometric calculations for pure 3D plane detection followed by validating their semantic category using point cloud data from RGB-D cameras.
It has a parallel multi-thread architecture to precisely estimate poses and equations of all the planes detected in the environment, filters the ones forming the map structure using a panoptic segmentation validation, and keeps only the validated building components.
Incorporating the proposed method into a VSLAM framework confirmed that constraining the map with the detected environment-driven semantic elements can improve scene understanding and map reconstruction accuracy.
It can also ensure (re-)association of these detected components into a unified 3D scene graph, bridging the gap between geometric accuracy and semantic understanding.
Additionally, the pipeline allows for the detection of potential higher-level structural entities, such as rooms, by identifying the relationships between building components based on their layout.
\end{abstract}
\section{Introduction}
\label{sec_intro}

\begin{figure}[t]
    \centering
    \includegraphics[width=1.0\columnwidth]{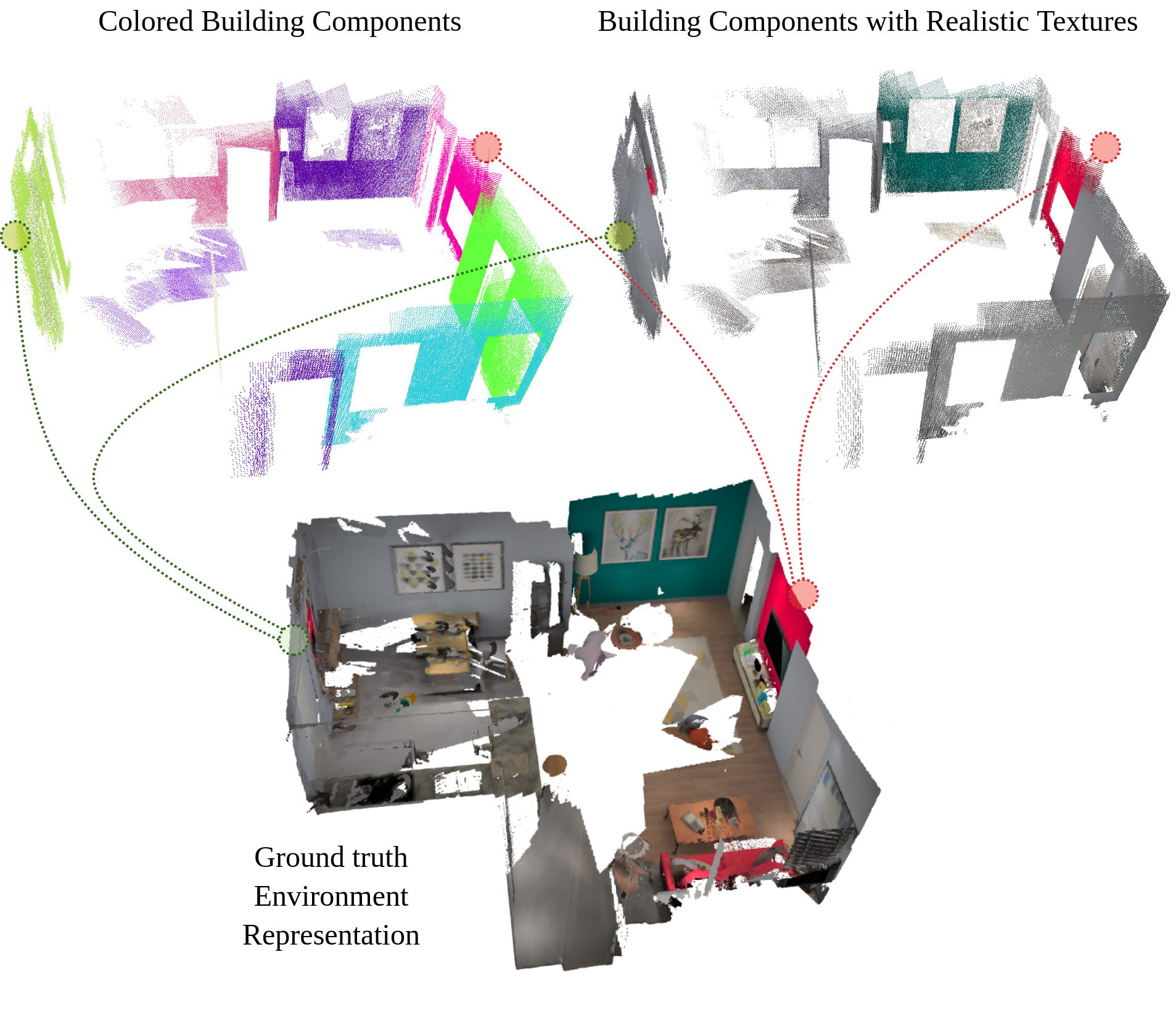}
    \caption{The 3D reconstructed map of an indoor environment using the proposed approach with \textbf{(left)} distinct color-coded building components for clear distinction and \textbf{(right)} the original RGB textures for a realistic representation.}
    \label{fig_overall}
\end{figure}

Employing the rich, dense data captured by sensors like \ac{LiDAR} or RGB-D cameras enables robots to effectively address challenges in environment understanding, reconstruction, and \ac{SLAM} \cite{hu2023robot}.
The point clouds generated by these sensors provide detailed spatial information, facilitating precise detection, localization, and 3D mapping of the environment elements essential for robotic perception \cite{li2023rgbd}.
In this context, RGB-D vision sensors offer a more comprehensive understanding of environments by fusing color and depth information, compensating for monocular cameras' lack of depth perception and LiDAR sensors' limited visual richness.

As vision sensors provide a versatile and cost-efficient strategy for map reconstruction, they established a unique extension of \ac{SLAM} algorithms titled \ac{VSLAM} \cite{pu2023visual}.
Computer vision methodologies are inevitable in \ac{VSLAM}, as they provide rich scene understanding and semantic object localization \cite{cai2024comprehensive, favorskaya2023deep}.
Such visual data can also form the scanned environment into optimizable 3D scene graphs and depict their representations in multi-level abstraction models \cite{zhang2024egosg, li2024rethinking}.
In both cases, providing integrated depth and visual information further enhances the robot's localization and mapping of the observed elements.

\begin{figure*}[t]
    \centering
    \includegraphics[width=1.0\textwidth]{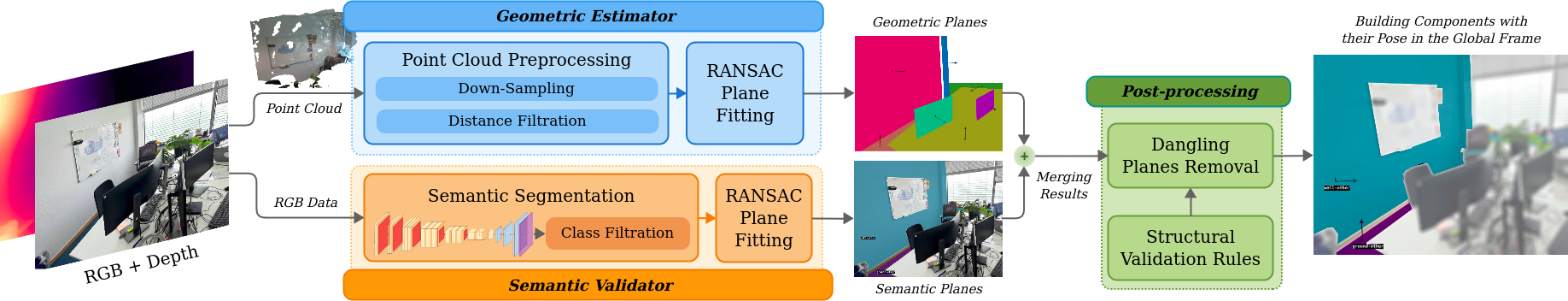}
    \caption{The order of steps to detect building components from RGB-D data using the proposed approach.}
    \label{fig_flowchart}
\end{figure*}

Given the advantages of point clouds generated by RGB-D cameras, this paper aims to employ combined depth and visual information for geometrical calculation and semantic confirmation, respectively.
The pipeline initiates with plane extraction using the depth data provided in the point clouds of a given frame and completes with semantic label augmentation through processing its visual data, resulting in rich information for building component localization.
Fig.~\ref{fig_overall} depicts the outcome of the proposed methodology in accurate mapping and scene understanding.
Accordingly, the primary contributions of the paper are summarized below:

\begin{itemize}
    \item A real-time pipeline for accurate recognition and localization of building-level semantic components, \ie wall and ground surfaces,
    \item A method for (re-)association of the detected building components and representing them in 3D scene graphs; and
    \item Incorporating the mentioned rich semantic entities to enhance the performance of \ac{VSLAM} through imposing environment-driven constraints.
\end{itemize}
\section{Related Works}
\label{sec_related}

The implication of localizing geometric and semantic entities is highly applicable in robotics and covers various fields of study.
In this regard, visual localization, feature matching, and pose estimation lay the foundation for navigation and mapping tasks, \eg \ac{VSLAM}.
Taira \etal \cite{taira2019right} introduced a methodology for proper pose verification of visited objects from different camera views by combining appearance, geometry, and semantics modalities.
Peng \etal \cite{peng20223d} presented a voxel-to-point algorithm to construct semantic and geometric features of given point clouds for outdoor environments.
In \cite{chen2023geosegnet}, semantic categories are assigned to point cloud data by extensively exploring geometric features extracted using a multi-geometry and boundary-guided encoder-decoder.
Despite their effectiveness, the introduced methods and many others \cite{shuai2023geometry, yang2023geometry, martens2023cross} have not integrated with \ac{SLAM} procedures for extensive map building and real-time localization.

On the other hand, enhancing \ac{VSLAM} applications with semantic and geometric constraints has shown potential for improving accuracy and scene understanding \cite{tourani2022visual}.
Accordingly, incorporating geometric components such as planes \cite{yan2024plpf, yang2023uplp} and lines \cite{zhang2024dynpl, dong2024plpd} helps reduce the environment's dimensionality, leading to more precise map reconstruction.
Alternatively, localizing and mapping semantic entities by estimating their poses using point cloud processing \cite{sandstrom2023point, drvslam, wei2023rgb, cong2024seg, zhang2023tsg, tang2024oms} or obtaining the pose information from fiducial markers attached to them \cite{semuco, vsgraphs1.0} offers accurate positioning and identification of critical environmental elements.
It can be seen that combining geometric and semantic constraints aids \ac{VSLAM} systems in achieving better scene comprehension and mapping of the detected entities.

The approach proposed in this paper pinpoints the gap between geometric object localization and semantic class verification within a unified framework, leading to accurate building component localization.
Filling this gap makes it possible to detect potential structural entities (\eg rooms) using building components (\ie wall and ground surfaces) regarding their layout and inner relationships, leading to the reconstruction of higher-accuracy maps.
\section{Proposed Method}
\label{sec_proposed}

Fig.~\ref{fig_flowchart} depicts the pipeline of the proposed methodology.
The process begins with a sequence of video frames captured by RGB-D cameras, providing colored visual data and point clouds enriched with depth information.
The pipeline can be optimized for \ac{VSLAM} by processing keyframes instead of every individual frame, reducing computational costs while maintaining the real-time performance for localization and mapping.
The point cloud and RGB data are passed to the "\textit{geometric estimator}" and "\textit{semantic validator}" modules, respectively, both running in parallel on separate threads.
The different stages of the pipeline are further detailed in the following subsections, where each module's functionality and part are explained:

\subsection{Geometric Estimator}
\label{sec_proposed_geo}

This module processes the point cloud data captured by the RGB-D sensor.
Its primary objective is to investigate the points within the cloud to detect and extract potential 3D planes (\ie planar regions) present in the environment.
Despite the point cloud's dense and rich spatial information, it may contain noisy or low-resolution points, negatively impacting subsequent processing steps.
Thus, the given point cloud \(\mathbf{P}=\{\mathbf{p}_i ~|~ i=1,2,…,n\}\) consisting of points \(\mathbf{p}=(x,y,z,r,g,b)\) undergoes a two-stage preprocessing procedure.
In the first step, down-sampling is applied to select \(\mathbf{P_d}=\{\mathbf{p}_i ~|~ i=1,2,…,m\}\), where \(n \leq m\) and \(\mathbf{P_d} \subset \mathbf{P}\).
Next, distance filtration is employed to remove outlier 3D points and keep only the ones that satisfy the condition \(\mathbf{P_f} = \{\mathbf{p}_i \in \mathbf{P_d} ~|~ {\theta}_{min} \leq \| \mathbf{p}_i \| \leq {\theta}_{max} \}\), where \({\theta}_{min}\) and \({\theta}_{max}\) are minimum and maximum desired depth range, respectively.
This results in the final preprocessed point cloud \(\mathbf{P_f}\).

The next step is to apply \ac{RANSAC} \cite{ransac} plane fitting algorithm on \(\mathbf{P_f}\) to detect potential planes with their geometric equations.
Hence, a set of random points \(\mathbf{{p_r}}_i \in \mathbf{P_f}\) are iteratively selected to calculate the normal vectors \(\mathbf{n}_j \in \mathbb{R}^3\) (\(j \in \mathbb{N}\)) representing a set of planes \(\pi_j\) with the pre-defined distance \(d(\mathbf{{p_r}}_i, \pi_j) \leq \epsilon\), where \(\epsilon\) is the inlier threshold.
The final output consists of the best-fitted geometric planes \(\mathbf{\Pi} = \{\pi_j~|~\pi_j \in \mathbb{R}^3, j \in \mathbb{N}\}\), representing all objects with planar surfaces detected from the filtered point cloud \(\mathbf{P_f}\).
It is essential to mention that the detected planes may correspond to various objects, such as tables, floors, screens, \etc (as shown in Fig.~\ref{fig_flowchart}), many of which are irrelevant for detecting building components required for understanding the environment's layout.
\subsection{Semantic Validator}
\label{sec_proposed_semantic}

In parallel to the \textit{geometric estimator}, RGB frames are passed into the \textit{semantic validator} module to extract semantic information, explicitly pivotal building components, including walls and ground surfaces.
This module applies semantic segmentation to label each pixel in the RGB data, ensuring that only relevant classes are kept while irrelevant entities are discarded.
Panoptic segmentation is preferred in this context due to its better occlusion handling, sharper boundary detection, and efficient distinguishing among objects within the same class.
Accordingly, the proposed method employs real-time panoptic semantic segmentation frameworks, including PanopticFCN \cite{pFCN} and YOSO \cite{yoso}.

Each frame \(f \in \mathbf{F}\) passes through the semantic segmentation function \(S\), resulting in the semantic label map \(L=S(f)\), where \(L_{(x,y)}\) refers to the semantic class label of the pixel at \((x,y) \in \mathbb{N}^2\).
The next step is to apply semantic \textit{class filtration} to filter building components' semantic classes \(\mathbf{\Psi}\) (\ie wall and ground surfaces) as follows:

\[
    {L_\Psi}_{(x,y)} = 
    \begin{cases} 
        L_{(x,y)} & \text{if } L_{(x,y)} \in \Psi \\
        \textit{null} & \textit{otherwise}
    \end{cases}
\]

\noindent where \(L_\Psi\) represents the filtered semantic label map derived from the input RGB frame, containing only the pixels corresponding to the building components specified in \(\mathbf{\Psi}\).
The uncertainty parameter \(\lambda_{(x,y)} \in \mathbb{R}\) for each pixel \({L_\Psi}_{(x,y)}\) is also stored for potential classification errors.

Since the semantic map \(L_\Psi\) may contain noisy and incomplete segments, it requires further refinement.
Therefore, another pass of \ac{RANSAC} is applied to \(L_\Psi\) to filter out noisy and incomplete parts, considering the uncertainty parameter \(\lambda\).
The outcome is referred to as \(\mathbf{\Pi}_\Psi = \{{\pi_\psi}_k~|~{\pi_\psi}_k \in \mathbb{R}^3, k \in \mathbb{N}\}\) and represents all building component objects derived from the \(L_\Psi\).
It should be noted that \(\mathbf{\Pi}_\Psi\) may suffer from sparsity due to segmentation resolution and occlusion constraints, making its geometric equations inconsistent.
An instance of the extracted labeled planar entities using the \textit{semantic validator} module is shown in Fig.~\ref{fig_flowchart}.
\subsection{Geometric-Semantic Fusion}
\label{sec_proposed_fusion}

The geometric planes \(\mathbf{\Pi}\) estimated by the \textit{geometric estimator} are more trustworthy due to being derived from dense point cloud data, delivering accurate geometric analyses.
However, these planes lack semantic information and can belong to any semantic entity with planar surfaces.
Conversely, the planes \(\mathbf{\Pi}_\Psi\) determined by the \textit{semantic validator} contain semantic labels belonging to specific building components, but they lack precise geometric equations.
Hence, the final stage of the pipeline involves associating \(\mathbf{\Pi}\) with \(\mathbf{\Pi}_\Psi\) planes belonging to frame \(f_t\) captured at time \(t\), enabling the validation of the geometric planes and assigning accurate semantic labels to them.
The essence of \(\mathbf{\Pi}_\Psi\) planes in the fusion stage is to validate the semantic labeling of \(\mathbf{\Pi}\) planes after matching, ensuring that the matched planes possess accurate geometric properties and are correctly identified as pre-defined building elements.

Considering geometric plane \(\pi_i \in \mathbf{\Pi}\) and semantic plane \({\pi_\psi}_j \in \mathbf{\Pi}_\Psi\), the association process is defined as follows:

\[
    \pi_i = 
    \begin{cases} 
        \pi_{\psi_j} & \text{if } \textit{match}(\pi_i, \pi_{\psi_j}) \geq \epsilon \\
        \textit{null} & \textit{otherwise}
    \end{cases}
\]

\noindent where \(\epsilon\) is a predefined threshold for a valid match with the most prominent inlier matches.
The matched (validated) planes are stored in \(\mathbf{\Pi}_\kappa \subset \mathbf{\Pi}\), which contain geometrically accurate and semantically meaningful planes.

The final step is applying post-processing algorithms on \(\mathbf{\Pi}_\kappa\).
Accordingly, the dangling removal function \(\mathbf{\Pi}_\kappa=D(\mathbf{\Pi}_\kappa)\) omits any planes of \(\mathbf{\Pi}_\kappa\) that do not correspond to wall or ground surfaces.
The remaining valid planes are passed through a structural validation function \(\mathbf{\Pi}_\kappa=V(\mathbf{\Pi}_\kappa)\), ensuring that they satisfy the necessary geometric constraints for the environment.
For instance, ground planes are constrained to be horizontal in the global reference frame, while wall planes must be vertical and perpendicular to the ground surfaces.
Detecting such building components prepares the ground for identifying high-level structural-semantic elements, such as rooms and corridors, defined by the topological associations among the mentioned elements.
\section{Evaluation}
\label{sec_evaluation}

This section describes the two assessments conducted to evaluate the proposed methodology's accuracy and usefulness.
One experiment analyzes the correct recognition of building components in the environment, while the second confirms its impact on map reconstruction by integrating with a \ac{VSLAM} framework.

\subsection{Evaluation Setup}
\label{sec_eval_setup}

The evaluations were conducted on a system equipped with an $11^{th}$ Gen. Intel® Core™ i9-11950H processor running at 2.60 GHz, an NVIDIA T600 Mobile GPU with 4GB of GDDR6 memory, and 32GB of RAM.
The employed datasets were ICL \cite{dataset_icl}, ScanNet \cite{dataset_scannet}, and an in-house dataset collected by the authors in real-world environments.
Table~\ref{tbl_dataset} presents the characteristics of the mentioned datasets used for evaluation.

\begin{table}[t]
    \centering
    \caption{The list of datasets used for evaluations.}
    \begin{tabular}{lcccccl}
        \toprule
        \textbf{\textit{Dataset}} & \textbf{\textit{\#Instances}} & \textbf{\textit{Duration}} & \textbf{\textit{Texturing}} \\
        \midrule
            \textit{ICL} \cite{dataset_icl} & 09 & 11m 26s & photo-realistic \\
            \textit{ScanNet} \cite{dataset_scannet} & 06 & 04m 32s & real \\
            \textit{In-house (collected)} & 07 & 41m 36s & real \\
            \midrule
            \textbf{Total} & \textbf{22} & \textbf{57m 34s} & \\
        \bottomrule
    \end{tabular}
    \label{tbl_dataset}
\end{table}

\begin{figure}[t]
    \centering
    \includegraphics[width=0.85\columnwidth]{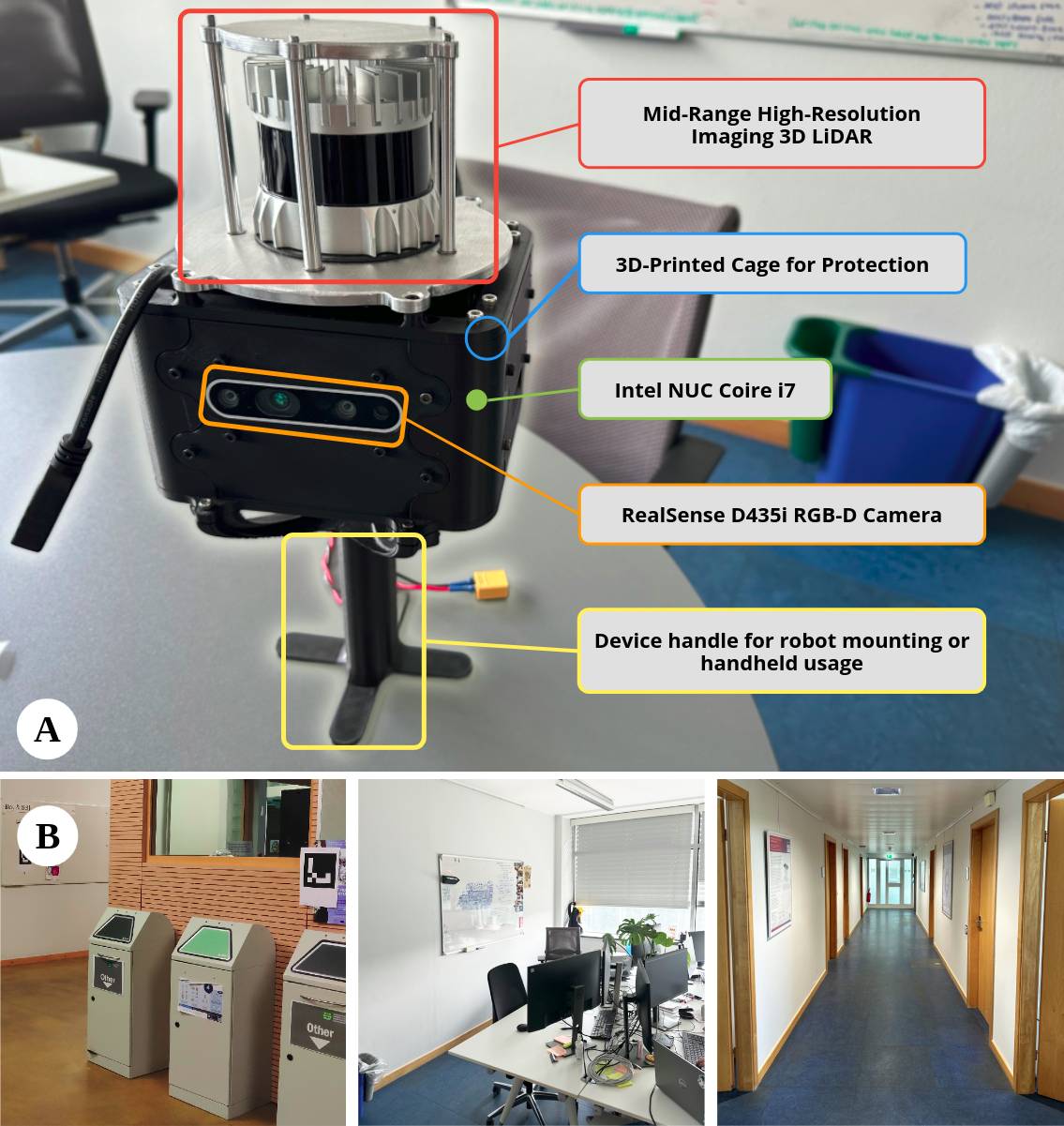}
    \caption{The in-house data collection setup \textbf{(A)} and some instances of the collected dataset \textbf{(B)}.}
    \label{fig_dataset}
\end{figure}

It should be noted that the in-house dataset contains sequences recorded in various indoor setups using a designed handheld/mountable device equipped with an \textit{Intel® RealSense™ Depth Camera D435i} and an \textit{Ouster OS1 Mid-Range High-Resolution Imaging \ac{LiDAR}} with the precision of $0.3-1m:\pm0.7cm$.
The \ac{LiDAR} data is primarily used as ground truth.
Fig.~\ref{fig_dataset} depicts the device employed for collecting data and some instances of our in-house dataset.
\subsection{Map Reconstruction Performance}
\label{sec_eval_slam}

To showcase the effectiveness of the proposed approach within a \ac{VSLAM} framework, we integrated our pipeline equipped with YOSO \cite{yoso} semantic segmentation into \orb~\cite{orbslam3}.
This integration detects and maps the defined building components (in this work wall and ground surfaces), adding additional constraints while preserving the system's real-time performance.
The evaluation results are presented in Table~\ref{tbl_slam_eval}, where the baseline (\orb) and its pipeline-integrated version were executed across eight iterations on the datasets to ensure reliable outcomes.
The employed evaluation metric is \ac{RMSE} in \textit{meters}.

\begin{table}[t]
    \centering
    \caption{Evaluation results on the various dataset using \acf{RMSE} in \textit{meters}. The best results are highlighted in bold. Numbers in brackets indicate the percentage of the traversed trajectory before tracking failure.}
    \begin{tabular}{ll|ccccc|cc|c}
        \toprule
         & &  \multicolumn{2}{c}{\textbf{Methodologies}} & \textbf{Diff.(\%)} \\
        \midrule
        \textbf{Dataset} & \textbf{Sequence} & \textit{\orb} & \textit{Proposed} \\
        \midrule
            ICL \cite{dataset_icl} & deer-gr     & 0.008~\textsuperscript{\tiny [58\%]} & \textbf{0.007}~\textsuperscript{\tiny [58\%]}& \cellcolor{greend}{\scriptsize +12.5\%} \\
                                & deer-wh     & 0.068 ~\textsuperscript{\tiny [53\%]} & 0.068~\textsuperscript{\tiny [53\%]} & \cellcolor{yellow}{\scriptsize 0.0\%} \\
                                & deer-w      & 0.098 ~\textsuperscript{\tiny [55\%]} & \textbf{0.094}~\textsuperscript{\tiny [61\%]} & \cellcolor{greend}{\scriptsize +4.1\%} \\
                                & deer-r      & 0.065 ~\textsuperscript{\tiny [64\%]} & \textbf{0.064}~\textsuperscript{\tiny [64\%]} & \cellcolor{greenl}{\scriptsize +1.5\%} \\
                                & deer-mavf   & 0.027 ~\textsuperscript{\tiny [cmp.]} & \textbf{0.026}~\textsuperscript{\tiny [cmp.]} & \cellcolor{greend}{\scriptsize +3.7\%} \\
                                & diam-gr     & 0.097 ~\textsuperscript{\tiny [cmp.]} & \textbf{0.068}~\textsuperscript{\tiny [cmp.]}  & \cellcolor{greend}{\scriptsize +29.9\%} \\
                                & diam-wh    & 0.041 ~\textsuperscript{\tiny [92\%]} & \textbf{0.029}~\textsuperscript{\tiny [cmp.]} & \cellcolor{greend}{\scriptsize +29.3\%} \\
                                & diam-mavs  & \textbf{0.017}~\textsuperscript{\tiny [cmp.]} &  0.018~\textsuperscript{\tiny [cmp.]} & \cellcolor{red}{\scriptsize -5.9\%} \\
                                & diam-mavf  & 0.027~\textsuperscript{\tiny [cmp.]} & \textbf{0.026}~\textsuperscript{\tiny [cmp.]} & \cellcolor{greend}{\scriptsize +3.7\%} \\
            \midrule
            ScanNet  & seq-0041-01  & 0.141~\textsuperscript{\tiny [cmp.]} & \textbf{0.136}~\textsuperscript{\tiny [95\%]} & \cellcolor{greend}{\scriptsize +3.5\%} \\
            \cite{dataset_scannet}  & seq-0103-01    & 0.077~\textsuperscript{\tiny [39\%]} & \textbf{0.076}~\textsuperscript{\tiny [39\%]} & \cellcolor{greenl}{\scriptsize +1.3\%} \\
                                    & seq-0200-00 & \textbf{0.047}~\textsuperscript{\tiny [61\%]} & 0.052~\textsuperscript{\tiny [61\%]} & \cellcolor{red}{\scriptsize -10.6\%} \\
                                    & seq-0560-00 & 0.090~\textsuperscript{\tiny [94\%]} & \textbf{0.092}~\textsuperscript{\tiny [cmp.]} & \cellcolor{orange}{\scriptsize -2.2\%} \\
                                    & seq-0614-01 & \textbf{0.126}~\textsuperscript{\tiny [72\%]} & 0.132~\textsuperscript{\tiny [69\%]} & \cellcolor{red}{\scriptsize -4.8\%} \\
                                    & seq-0626-00 & 0.095~\textsuperscript{\tiny [cmp.]} & \textbf{0.094}~\textsuperscript{\tiny [cmp.]} & \cellcolor{greenl}{\scriptsize +1.1\%} \\
            \midrule
            Collected   & seq-1rm & 0.063~\textsuperscript{\tiny [cmp.]} & 0.063~\textsuperscript{\tiny [cmp.]} & \cellcolor{yellow}{\scriptsize 0.0\%} \\
            (in-house)  & seq-4rm-2crr & 0.432~\textsuperscript{\tiny [cmp.]} & \textbf{0.418}~\textsuperscript{\tiny [cmp.]} & \cellcolor{greend}{\scriptsize +3.2\%} \\
                        & seq-nested & \textbf{0.163}~\textsuperscript{\tiny [88\%]} & 0.169~\textsuperscript{\tiny [98\%]} & \cellcolor{red}{\scriptsize -3.7\%} \\
                        & seq-1rm-1crr & 0.278~\textsuperscript{\tiny [cmp.]} & \textbf{0.262}~\textsuperscript{\tiny [cmp.]} & \cellcolor{greend}{\scriptsize +5.8\%} \\
                        & seq-4crr & 0.285~\textsuperscript{\tiny [75\%]} & \textbf{0.269}~\textsuperscript{\tiny [75\%]} & \cellcolor{greend}{\scriptsize +5.6\%} \\
                        & seq-2rm-1crr & \textbf{0.125}~\textsuperscript{\tiny [cmp.]} & 0.126~\textsuperscript{\tiny [cmp.]} & \cellcolor{orange}{\scriptsize -0.8\%} \\
                        & seq-lab & \textbf{0.376}~\textsuperscript{\tiny [99\%]} & 0.377~\textsuperscript{\tiny [cmp.]} & \cellcolor{orange}{\scriptsize -0.3\%} \\
        \bottomrule
    \end{tabular}
    \label{tbl_slam_eval}
\end{table}

\begin{figure}[t]
     \centering
     \begin{subfigure}[t]{0.23\textwidth}
         \centering
         \includegraphics[width=0.98\columnwidth]{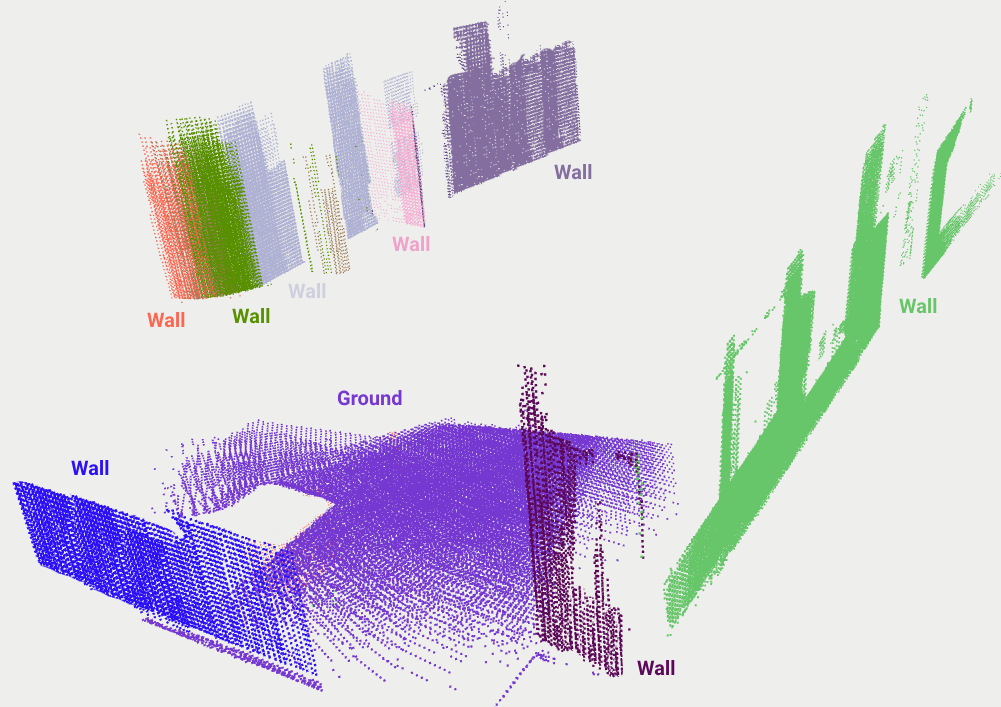}
         \caption{\textit{diam-gr} of ICL \cite{dataset_icl}}
         \label{fig_eval_slam_icl}
     \end{subfigure}
     \begin{subfigure}[t]{0.23\textwidth}
         \centering
         \includegraphics[width=0.98\columnwidth]{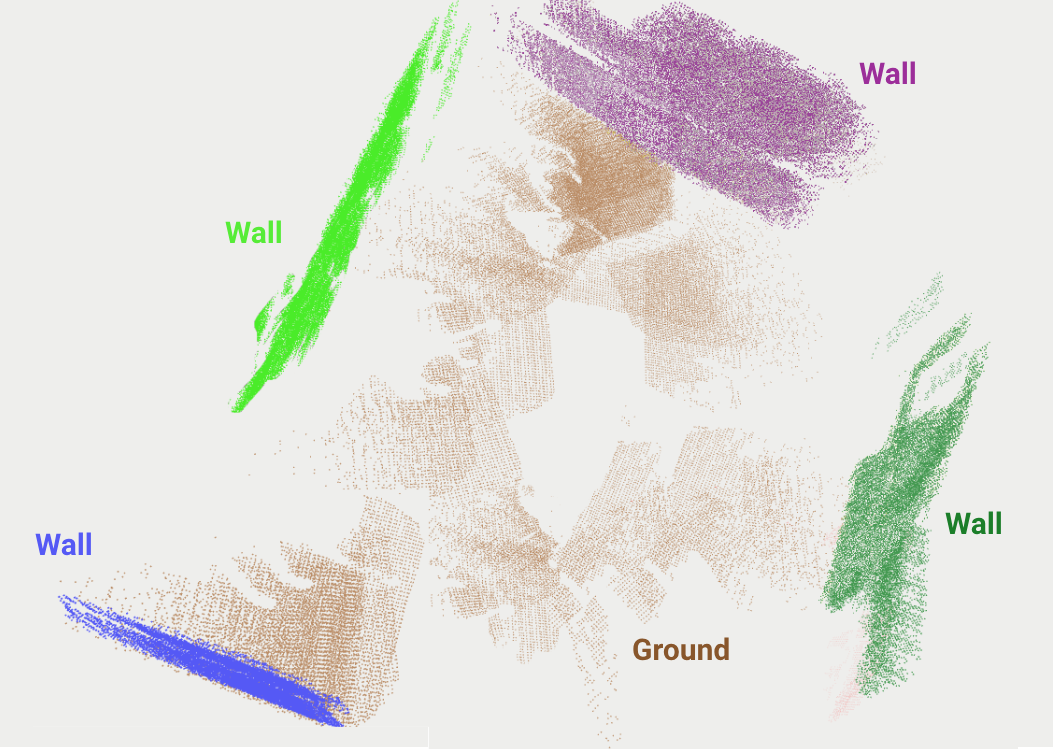}
         \caption{\textit{0041-01} of ScanNet \cite{dataset_scannet}}
         \label{fig_eval_scan}
     \end{subfigure}
     \begin{subfigure}[t]{0.23\textwidth}
         \centering
         \includegraphics[width=0.98\columnwidth]{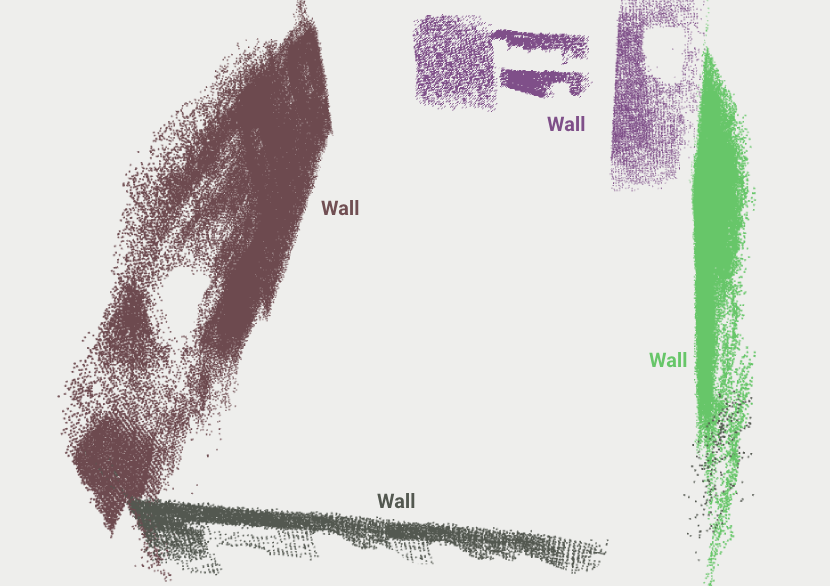}
         \caption{\textit{seq-1rm} of in-house}
         \label{fig_eval_slam_lux1}
     \end{subfigure}
     \begin{subfigure}[t]{0.23\textwidth}
         \centering
         \includegraphics[width=0.98\columnwidth]{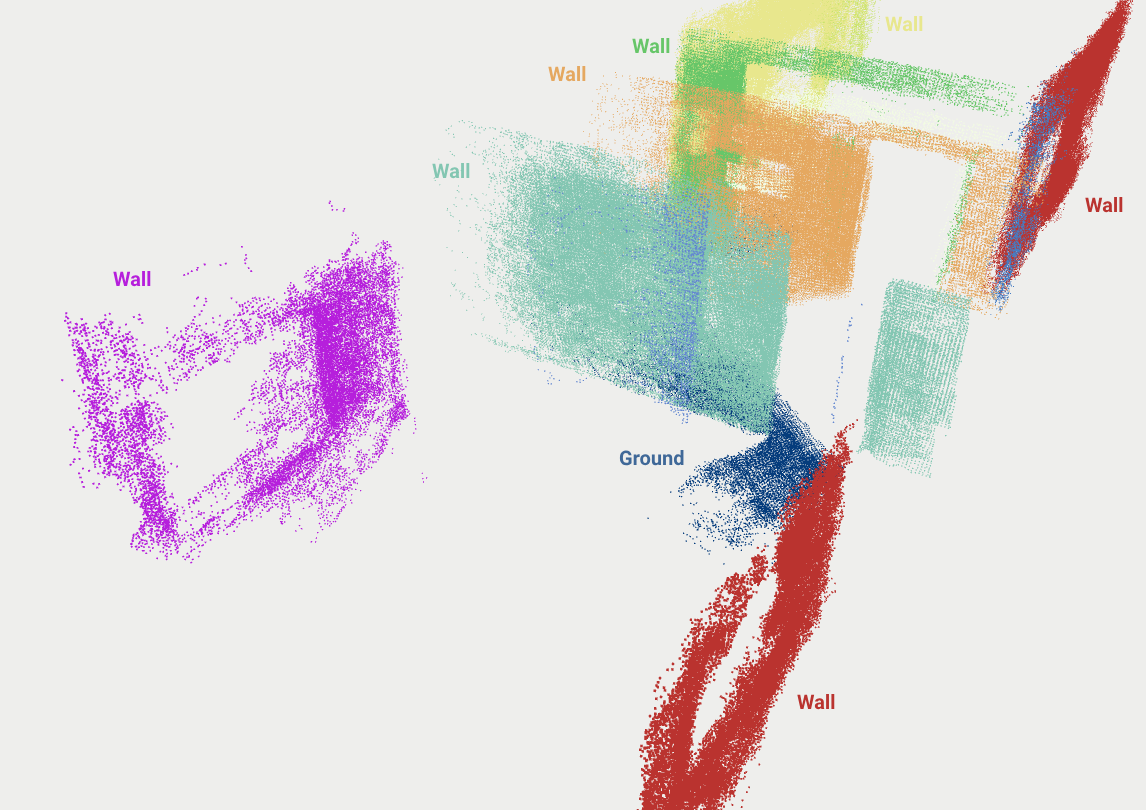}
         \caption{\textit{seq-2rm-1crr} of in-house}
         \label{fig_eval_lux2}
     \end{subfigure}
     \caption{Reconstructed maps presented in 3D scene graphs enriched with recognized building components in some dataset instances. Labels have been manually added to enhance clarity.}
     \label{fig_eval_slam}
\end{figure}

As shown in Table~\ref{tbl_slam_eval}, the pipeline-integrated version of \orb~ outperforms the baseline across various environment layouts in most of the cases.
This improvement is primarily due to incorporating environment-specific constraints derived from accurately detecting building components, including wall and ground surfaces.
With such constraints, the system achieves a more structured and informed reconstructed map, reducing trajectory estimation errors and better aligning with real-world features.
In the ICL dataset, where the environment is relatively small, the most significant improvements -around 30\%- occurred in \textit{diam-gr} and \textit{diam-wh}, with reductions of \textit{2.9 cm} and \textit{1.2 cm} in \ac{RMSE}, respectively.
Sequence \textit{seq-0200-00} in the ScanNet dataset leads to a higher trajectory error of \textit{0.6 cm}, presumably due to errors in associating building components after getting lost and incomplete map reconstruction affecting overall accuracy.
In the collected real-world in-house dataset where the sequences are longer, the best results appear in \textit{seq-1rm-1crr} (one office connected to a corridor) and \textit{seq-4crr} (four connected corridors) with \textit{5.8\%} and \textit{5.6\%} improvements, respectively.
Some instances of the reconstructed maps presented in 3D scene graphs are depicted in Fig.~\ref{fig_eval_slam}.

\subsection{Building Component Recognition Accuracy}
\label{sec_eval_rec}

Table~\ref{tbl_eval_rec} presents the performance of the proposed approach in accurately detecting building components across different datasets.
For evaluation, we calculated the values of \ac{TP}, \ac{FP}, and \ac{FN} for each dataset instance.
In this context, a \ac{TP} represents a building component that exists in the environment and is correctly detected by the method.
A \ac{FN} occurs when a building component is incorrectly detected where none exists, while an \ac{FP} indicates a failure to detect an existing component.
Accordingly, the proposed method's performance was assessed using object-level precision, recall, and the F1-Score calculations.
While precision measures the ratio of correctly detected components relative to all detected ones, recall reflects the method's ability to detect all existing building components, making it possible to provide a balanced metric that covers both measurements as follows:

\[ F1 = 2 \times \dfrac{precision \times recall}{precision + recall} \]

\begin{table}[t]
    \centering
    \scriptsize
    \caption{The accuracy of the proposed method in correctly detecting building components.}
    \begin{tabular}{ll|c|cccc}
        \toprule
         & & \textbf{Ground truth} & \multicolumn{3}{c}{\textbf{Methodologies}} \\
        \midrule
        \textbf{Dataset} & \textbf{Sequence} & \textit{\#Components} & \textit{Precision} & \textit{Recall} & \textit{F1-Score} \\
        \midrule
            ICL \cite{dataset_icl}  & deer-gr & 5 & \cellcolor{greenl}{0.80} & \cellcolor{greenl}{0.80} & \cellcolor{greenl}{0.80} \\
                                    & deer-wh & 10 & \cellcolor{greend}{1.00} & \cellcolor{greend}{1.00} & \cellcolor{greend}{1.00} \\
                                    & deer-w & 10 & \cellcolor{greend}{0.91} & \cellcolor{greend}{1.00} & \cellcolor{greend}{0.95} \\
                                    & deer-r & 10 & \cellcolor{greend}{1.00} & \cellcolor{greenl}{0.80} & \cellcolor{greenl}{0.89} \\
                                    & deer-mavf & 5 & \cellcolor{greenl}{0.80} & \cellcolor{greenl}{0.80} & \cellcolor{greenl}{0.80} \\
                                    & diam-gr & 6 & \cellcolor{orange}{0.75} & \cellcolor{greend}{1.00} & \cellcolor{greenl}{0.86} \\
                                    & diam-wh & 7 & \cellcolor{orange}{0.71} & \cellcolor{orange}{0.71} & \cellcolor{orange}{0.71} \\
                                    & diam-mavs & 4 & \cellcolor{greend}{1.00} & \cellcolor{red}{0.50} & \cellcolor{yellow}{0.67} \\
                                    & diam-mavf & 4 & \cellcolor{greend}{1.00} & \cellcolor{red}{0.50} & \cellcolor{yellow}{0.67} \\
            \midrule
            ScanNet  & seq-0041-01  & 5 & \cellcolor{greenl}{0.83} & \cellcolor{greend}{1.00} & \cellcolor{greend}{0.91} \\
            \cite{dataset_scannet}  & seq-0103-01 & 5 & \cellcolor{greend}{1.00} & \cellcolor{yellow}{0.60} & \cellcolor{orange}{0.75} \\
                                    & seq-0200-00 & 5 & \cellcolor{greend}{1.00} & \cellcolor{greend}{1.00} & \cellcolor{greend}{1.00} \\
                                    & seq-0560-00 & 5 & \cellcolor{greend}{1.00} & \cellcolor{greend}{1.00} & \cellcolor{greend}{1.00} \\
                                    & seq-0614-01 & 5 & \cellcolor{greend}{1.00} & \cellcolor{greend}{1.00} & \cellcolor{greend}{1.00} \\
                                    & seq-0626-00 & 5 & \cellcolor{greenl}{0.83} & \cellcolor{greend}{1.00} & \cellcolor{greend}{0.91} \\
            \midrule
            Collected   & seq-1rm  & 5 & \cellcolor{greenl}{0.80} & \cellcolor{greenl}{0.80} & \cellcolor{greenl}{0.80} \\
            (in-house)  & seq-4rm-2crr  & 20 & \cellcolor{greenl}{0.86} & \cellcolor{greend}{0.95} & \cellcolor{greend}{0.90} \\
                        & seq-nested  & 9 & \cellcolor{orange}{0.70} & \cellcolor{orange}{0.78} & \cellcolor{orange}{0.74} \\
                        & seq-1rm-1crr  & 7 & \cellcolor{greenl}{0.86} & \cellcolor{greenl}{0.86} & \cellcolor{greenl}{0.86} \\
                        & seq-4crr  & 18 & \cellcolor{greenl}{0.84} & \cellcolor{greenl}{0.89} & \cellcolor{greenl}{0.86} \\
                        & seq-2rm-1crr  & 7 & \cellcolor{orange}{0.75} & \cellcolor{red}{0.43} & \cellcolor{red}{0.55} \\
                        & seq-lab  & 23 & \cellcolor{orange}{0.79} & \cellcolor{greenl}{0.83} & \cellcolor{greenl}{0.81} \\
            \midrule
            \textbf{Total} & & \textbf{180} & \textbf{0.87} & \textbf{0.83} & \textbf{0.84} \\
        \bottomrule
    \end{tabular}
    \label{tbl_eval_rec}
\end{table}

According to Table~\ref{tbl_eval_rec}, the presented method functions satisfactorily in most scenarios, consistently acquiring high precision and recall over various datasets' instances.
The good results are attributed to the integration of geometric detection followed by semantic validation, improving the accuracy of structural elements in the environment.
Slight deviations in performance may occur depending on occlusions, the complexity of the environment, or incomplete data observation in the semantic maps.

\begin{figure}[t]
     \centering
     \begin{subfigure}[t]{0.23\textwidth}
         \centering
         \includegraphics[width=0.98\columnwidth]{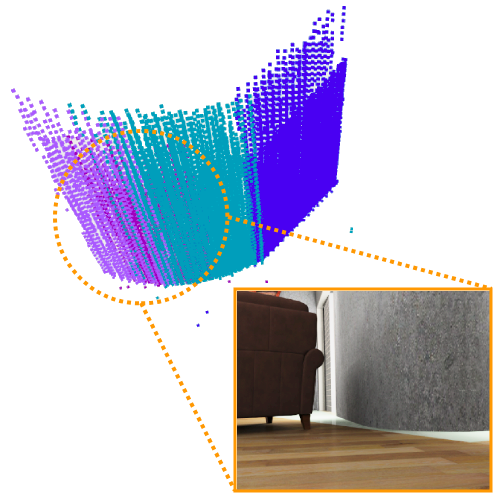}
         \caption{A curved wall in \textit{diam-wh}}
         \label{fig_eval_rec_curve}
     \end{subfigure}
     \begin{subfigure}[t]{0.23\textwidth}
         \centering
         \includegraphics[width=0.98\columnwidth]{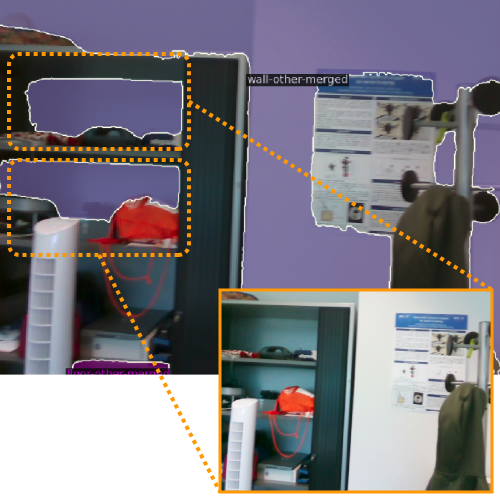}
         \caption{Segmentation inaccuracy}
         \label{fig_eval_rec_semseg}
     \end{subfigure}
     \hfill
     \begin{subfigure}[t]{0.23\textwidth}
         \centering
         \includegraphics[width=0.98\columnwidth]{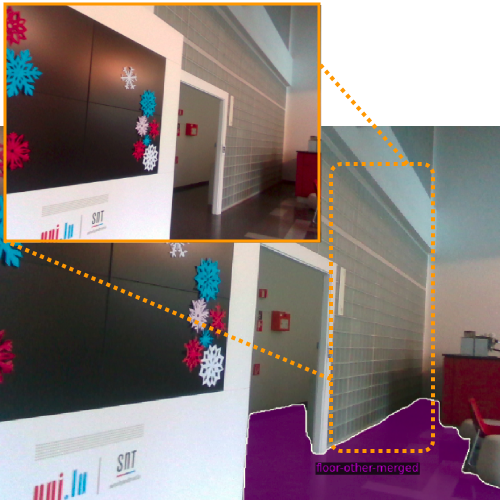}
         \caption{Glass block in \textit{seq-2rm-1crr}}
         \label{fig_eval_rec_glass}
     \end{subfigure}
     \begin{subfigure}[t]{0.23\textwidth}
         \centering
         \includegraphics[width=0.98\columnwidth]{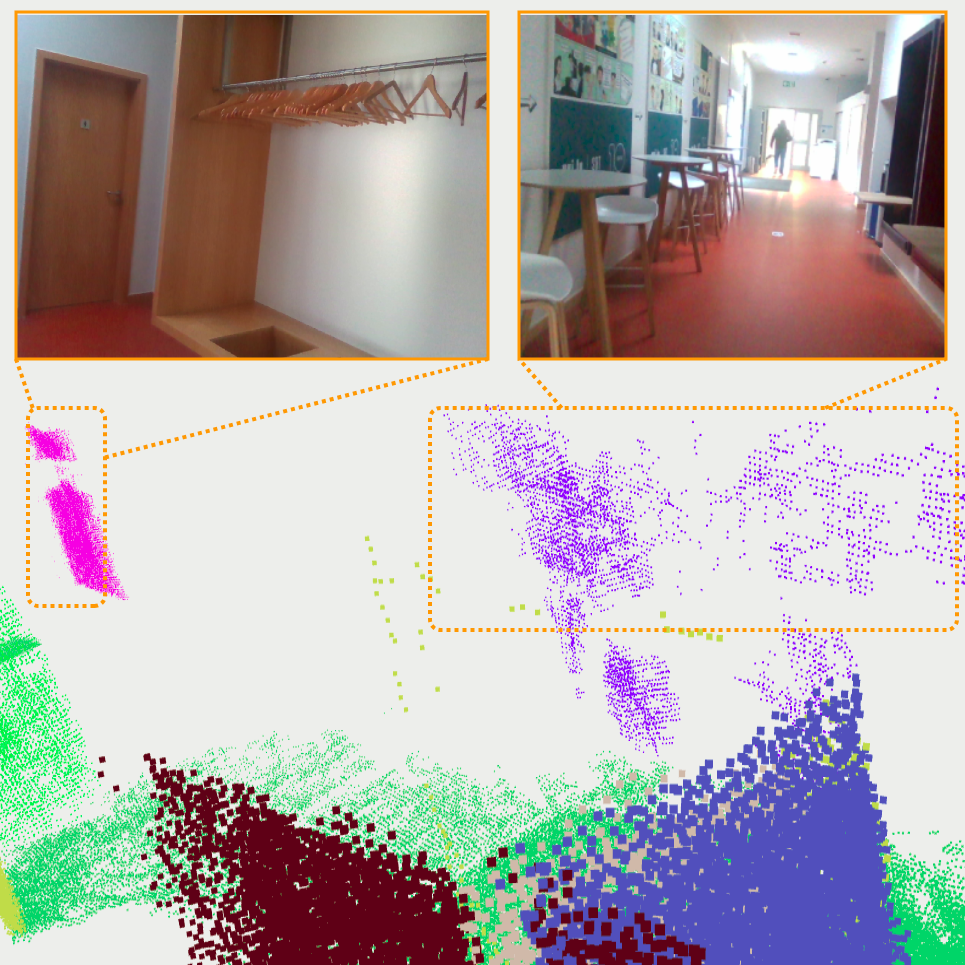}
         \caption{Association in \textit{seq-nested}}
         \label{fig_eval_rec_assoc}
     \end{subfigure}
     \caption{Edge-case scenarios posing challenges on the proposed method: \textbf{a)} multiple walls detected due to the curved shape of the wall, \textbf{b)} misclassification of some parts of the cupboard as a wall, \textbf{c)} ignoring glass blocks as walls, and \textbf{d)} the same walls are not associated due to pose error.}
     \label{fig_eval_rec}
\end{figure}

On the other hand, several factors can be attributed to the causes of inadequate precision or recall results in some cases.
A sample is an incorrect association between geometric and semantic entities due to fast camera motion during data acquisition, leading to tracking loss or visual feature mismatching.
Additionally, curved wall designs pose challenges on association and validation, as the method mainly relies on planar surfaces for building component detection.
Other factors include inaccuracies in semantic segmentation for classifying texture-less flat surfaces (\eg cupboards and desk drawers) and the existence of walls with glass blocks in the environment.
Fig.~\ref{fig_eval_rec} depicts some of the mentioned limitations of the proposed method in challenging scenarios.

\section{Conclusions}
\label{sec_conclusions}

This paper presented a real-time pipeline for integrating geometrical planar surface detection and object-level semantic validation to detect building components (in this case, wall and ground surfaces) on RGB-D point cloud data.
The pipeline aims to accurately recognize entities critical for high-level scene understanding by applying 3D plane poses extracted from RANSAC-based plane fitting and keeping/discarding those planes after validating their semantic category using a panoptic segmentation.
Experimental results revealed that the proposed method could improve scene understanding and representation in 3D scene graphs, besides enhancing the accuracy of a visual SLAM framework by imposing new environment-driven constraints.

As a continuance of the work, we are currently working to incorporate the proposed method into a real-time \ac{VSLAM}, empowering it to detect high-level structural-semantic entities alongside performing semantic loop closing.


\bibliographystyle{IEEEtran}
\bibliography{root}

\end{document}